# Multidimensional Digital Filters for Point-Target Detection in Cluttered Infrared Scenes

**Hugh L. Kennedy** [a]

[a] University of South Australia, Defence and Systems Institute (DASI), School of Engineering, Mawson Lakes Boulevard, Adelaide, Australia, 5095

**Abstract**. A 3-D spatiotemporal prediction-error filter (PEF), is used to enhance foreground/background contrast in (real and simulated) sensor image sequences. Relative velocity is utilized to extract point-targets that would otherwise be indistinguishable on spatial frequency alone. An optical-flow field is generated using local estimates of the 3-D autocorrelation function via the application of the fast Fourier transform (FFT) and inverse FFT. Velocity estimates are then used to 'tune in' a background-whitening PEF that is matched to the motion and texture of the local background. Finite-impulse-response (FIR) filters are designed and implemented in the frequency domain. An analytical expression for the frequency response of velocity-tuned FIR filters, of odd or even dimension, with an arbitrary 'delay' in each dimension, is derived.

**Address all correspondence to:** Hugh L. Kennedy, University of South Australia, Defence and Systems Institute (DASI), School of Engineering, Mawson Lakes Boulevard, Adelaide, Australia, 5095; Tel: +61 8 8302 5591; Fax: +61 8 8302 5344; E-mail: hugh.kennedy@unisa.edu.au

## 1 Introduction

Early automatic detection of airborne targets at long ranges that are set against textured backgrounds (e.g. cloud, sea, terrain or foliage) using infrared sensors, is a problem that continues to attract the attention of practitioners and theorists alike. Despite recent advances in thermal-imaging and data-processing technologies, it is almost certain that operators, who rely on such systems to successfully complete their missions, will always demand improved performance.

Simply applying a threshold to extract possible target-detections for a tracker is an unsatisfactory solution due to the high number of correlated clutter-detections produced by background features. On the one hand, consecutive frame differencing or the application of a one-dimensional (1-D) high-pass filter of low order in the temporal domain[1] is a very simple and effective approach in many situations (e.g. blue sky); however, this is likely to produce a high probability of false alarm for dynamic backgrounds and a low probability of detection for static targets. On the other hand, background estimation and subtraction algorithms or other high-pass filtering frameworks, operating in two-dimensions (2-D) on each frame in isolation – such as Wiener filters[2], least-mean-squares filters[3,4,5], top-hat transforms[6], moving average filters[7], median[7] and bilateral[3,7] filters – clearly do not suffer from these problems; however, the powerful discriminants of temporal coherence and disparity, which are essential cues in biological vision systems, are lost. Some methods attempt to solve this problem using one type of 1-D filter in the temporal dimension and a different type of 2-D filter in the spatial dimension[8].

Three-dimensional (3-D) filters provide a convenient mechanism for the integration of the spatial and temporal axes into a coherent framework and offer a wide range of design alternatives[9] – finite impulse response (FIR) or infinite impulse response (IIR), recursive or non recursive, with nominal pass-bands of arbitrary shape (e.g. plane, beam, wedge/fan[9,10,11], pyramid[12], cone[13], donut[14], etc.). While these filters have proven to be very effective in novel imaging, audio/acoustic and radio-frequency applications[9], they offer rapidly diminishing returns when they are applied to the problem of foreground enhancement and background cancellation in infrared sensors, because typical scenes of interest are highly non-stationary, due to object edges/boundaries for instance. Velocity-tuned filter-banks are another somewhat more computationally expensive 3-D solution to the problem of dim point-target detection[15,16,17,18]; however they are usually only applied after the background has been pre-'whitened'[2].

The method described in this paper aims for a compromise between the simplicity of 2-D spatial and 1-D temporal high-pass filters at one extreme and the complexity of optimal 3-D filters at the other. The 2-D moving-average prediction-error filter and the 1-D polynomial prediction-error filters could indeed be regarded as being limiting cases of the proposed approach. In the former case, only a direct-'current' (DC) spatial component with one-frame temporal support and wide-area spatial support is considered; whereas in the latter case a higher-order model is used with one-pixel spatial support and temporal support of many frames. In the approach described here, complex sinusoids are used instead of polynomials and the designer is free to choose both the extent (i.e. 'support') and model order in each dimension. As a *linear* model, there is some smearing/blurring of sharp edges, and as the spatial order is increased, this is replaced by damped 'ringing' phenomena.

The spatial pass-band of the prediction error filter (PEF) is simply defined using a rectangular grid of frequency *samples*. Velocity selective filters are then designed and an optical-flow field is used to select the most appropriate filter to apply. Velocity estimates are derived from the 3-D autocorrelation function which is computed efficiently using the 3-D fast-Fourier transform (FFT). The background subtraction filter is also applied in the frequency domain and non-linear-phase filters are derived. In the block-centric architecture employed here, this is mainly used to increase the number of pixels that can be processed with each FFT; however, in a sliding or recursive framework, this allows the phase delay of the filters to be tuned to yield the desired tradeoff between filter latency (which is not desirable in a closed-loop control application) and filter response (which is degraded as the latency decreases).

General closed-form expressions for the filter coefficients are derived in Sec. 2.1 along with an analytical expression for their frequency response; a description of the velocity estimation algorithm follows in Sec 2.2. Further implementation details are discussed in Sec. 3 then a frequency-domain realization is used to process synthetic data in Sec. 4 and real infrared data in Sec. 6. Issues associated with the exploitation of non-linear phase FIR filters are discussed in Sec. 5. The paper closes with some concluding remarks in Sec. 7.

## 2 Formulation

Use of a 3-D velocity-tuned filter, in principle, allows foreground and background signals that overlap substantially in spatial frequency to be resolved on spatiotemporal frequency separation brought about by apparent motion differences. It is assumed that the structured/textured background may experience spatially non-uniform motion, so that image registration and simple frame differencing approaches are not applicable.

Formulating the background/foreground separation problem as a prediction-error problem, where the background is ideally reduced to a white-noise field, permits the use of a simple peak-detection stage, followed by Bayesian tracking algorithm to automatically initiate and confirm target tracks, to refine state estimates and to maintain the continuity of target identity in a measurement space populated by uniformly distributed clutter. The 3-D FIR prediction-error filters are designed using a non-iterative frequency-sampling approach. The design process is effectively partitioned into two stages: the 'analysis' stage involves the *estimation* (in a least-squares-sense) of the background model parameters, using 'noisy' data within a 3-D analysis window; while the 'synthesis' stage *applies* the model to estimate the value of the background at a specified synthesis sample, which for best results and minimal phase non-linearity, is located near the centre of the analysis window. The two stages are combined and applied using a single 3-D operator in either the sample or frequency domain. The analysis provided in this Section is the main contribution of this paper. Many of the relationships described below may also be applied to other filter realizations, for example: frequency domain or sample domain, IIR or FIR, recursive or non-recursive.

### *2.1 Filter Design*

The proposed method is most effective in situations where the spatial structure or 'texture' of the background may be expressed using just a few spatially-extended low-frequency sinusoids with parameters (phase and amplitude) that vary only slowly in space and time (i.e. approximately locally stationary). The derivation begins with a 2-D spatial model of the background where an $M_x \times M_y$ analysis window is used to process an $N_x \times N_y$ image. The window and image are indexed, in opposite directions, using $[m_x, m_y]$ and $[n_x, n_y]$, respectively; thus the origin of the window $\boldsymbol{m} = [0,0]$, is at $\boldsymbol{n} = [n_x, n_y]$. The intensity $I$, of the (monochrome) pixels within the analysis window in a given frame, is modeled as a linear combination of complex sinusoidal basis functions with additive noise

$$I(n_x - m_x, n_y - m_y) = \sum_{k_y=-B_y}^{+B_y} \sum_{k_x=-B_x}^{+B_x} \beta(k_x, k_y) F^*\left(m_x, m_y; \frac{k_x}{M_x}, \frac{k_y}{M_y}\right) + \varepsilon \quad (1)$$

where the asterisk superscript denotes complex conjugation, the noise is distributed as a zero-mean Gaussian variable $\varepsilon \sim \mathcal{N}(0, \sigma^2)$ and

$$F(m_x, m_y; f_x, f_y) = \frac{1}{\sqrt{M_x M_y}} e^{j2\pi(f_x m_x + f_y m_y)} \quad (2)$$

are the 2-D sinusoidal components representing the band-limited background, with the number of components less than the window length in each dimension ($W_x = 2B_x + 1$, $W_x < M_x$ and $W_y = 2B_y + 1$, $W_y < M_y$). According to this simple model, the analysis window dimensions are an integer multiple of the component wavelengths, therefore the component indices $[k_x, k_y]$ denote the number of completed cycles within the analysis window (i.e. the 'wave number'), thus the components form an orthonormal basis set, permitting the Maximum Likelihood Estimate (MLE) of the component coefficients to be determined using

$$\hat{\beta}(k_x, k_y) = \sum_{m_y=0}^{M_y-1} \sum_{m_x=0}^{M_x-1} F(m_x, m_y; k_x, k_y) I(n_x - m_x, n_y - m_y) \tag{3}$$

where the 'hat' accent denotes an estimated quantity. When the background is in motion with velocity $\boldsymbol{v} = [v_x, v_y]$, the 2-D spatial components are 'tilted' in the 3-D frequency space[15]. They now have a non-zero normalized frequency of $f_z$ (in units of cycles per frame) in the temporal dimension $z$. Thus the background intensity $I$, at $\boldsymbol{m} = [m_x, m_y, m_z]$, within a $M_x \times M_y \times M_z$ window, with its origin at $\boldsymbol{n} = [n_x, n_y, n_z]$ is

$$I(\boldsymbol{n} - \boldsymbol{m}) = \sum_{k_y=-B_y}^{B_y} \sum_{k_x=-B_x}^{B_x} \beta(k_x, k_y) G^* \left(\boldsymbol{m}; \frac{k_x}{M_x}, \frac{k_y}{M_y}, \boldsymbol{v}\right) + \varepsilon \tag{4}$$

with 3-D sinusoidal components

$$G(\boldsymbol{m}; f_x, f_y, \boldsymbol{v}) = \frac{1}{\sqrt{M_x M_y M_z}} e^{j2\pi(f_x m_x + f_y m_y + f_z m_z)} \tag{5a}$$

where

$$f_z = -v_x f_x - v_y f_y . \tag{5b}$$

This simple model does have some obvious limitations. It assumes that the background only contains the specified frequency components within the analysis window, which is unlikely in most scenes of interest. Even when this condition is satisfied however, the model assumes that there is no non-linear dispersion within the window, i.e. that all components move with the same group velocity, which will not be the case if objects in the background are at different ranges from the sensor or experience non-rigid motion.

The sinusoidal basis retains its orthonormality after 'rotation', therefore the component coefficients (or model parameters) are estimated using the 'analysis' equation

$$\begin{aligned} \hat{\beta}(k_x, k_y) &= \sum_{m_z=0}^{M_z-1} \sum_{m_y=0}^{M_y-1} \sum_{m_x=0}^{M_x-1} G\left(\boldsymbol{m}; \frac{k_x}{M_x}, \frac{k_y}{M_y}, \boldsymbol{v}\right) I(\boldsymbol{n} - \boldsymbol{m}) \\ &= \sum_{\boldsymbol{m}=0}^{\boldsymbol{M}-1} G\left(\boldsymbol{m}; \frac{k_x}{M_x}, \frac{k_y}{M_y}, \boldsymbol{v}\right) I(\boldsymbol{n} - \boldsymbol{m}) . \end{aligned} \tag{6}$$

As the incoming frames are stored in a sliding window of length $M_z$, it is convenient to index the data within the analysis window and the data stream in opposite directions, so that $m_z = 0$ always corresponds to the most recent frame. For consistency and conformity with convention,

‘delay indexing’ is also applied in the spatial dimensions even though it is not really necessary because the spatial dimensions are of finite extent and all pixels in a frame, for all intents and purposes, arrive simultaneously; therefore non-causal indexing and filtering is feasible. The discretized spatio-temporal data (voxels) will be collectively referred to as ‘samples’ because it is not necessary to discriminate between spatial data (or pixels) and temporal data (or frames) in the treatment that follows. Note also that to simplify notation, a single summation over the vector of $\boldsymbol{m}$ indices is used in Eq. 6 to represent the summation over all $m_x$, $m_y$ and $m_z$ indices from [0,0,0] to $[M_x - 1, M_y - 1, M_z - 1]$; this convention is used throughout this Section.

With the background model-parameters estimated, the model may be evaluated (i.e. ‘synthesized’) at a sample within the analysis window (smoothing), in between samples (interpolation) or outside the analysis window (extrapolation) to give the noise-free estimate of the background intensity, $\hat{I}$. Substitution of Eq. 6 into Eq. 4 yields

$$\hat{I}(\boldsymbol{n} - \acute{\boldsymbol{m}}) = \sum_{k_y=-B_y}^{+B_y} \sum_{k_x=-B_x}^{+B_x} \sum_{\boldsymbol{m}=0}^{\boldsymbol{M}-1} G^*\left(\acute{\boldsymbol{m}}; \frac{k_x}{M_x}, \frac{k_y}{M_y}, \boldsymbol{v}\right) G\left(\boldsymbol{m}; \frac{k_x}{M_x}, \frac{k_y}{M_y}, \boldsymbol{v}\right) I(\boldsymbol{n} - \boldsymbol{m}) \quad (7)$$

where $G^*\left(\acute{\boldsymbol{m}}; \frac{k_x}{M_x}, \frac{k_y}{M_y}, \boldsymbol{v}\right)$ is the complex conjugate of $G\left(\boldsymbol{m}; \frac{k_x}{M_x}, \frac{k_y}{M_y}, \boldsymbol{v}\right)$, evaluated at $\acute{\boldsymbol{m}} = [\acute{m}_x, \acute{m}_y, \acute{m}_z]$. As the summation over the components is not data dependent, ‘analysis’ and ‘synthesis’ operations may be combined, therefore Eq. 7 reduces to

$$\hat{I}(\boldsymbol{n} - \acute{\boldsymbol{m}}) = \sum_{\boldsymbol{m}=0}^{\boldsymbol{M}-1} H(\boldsymbol{m}; \acute{\boldsymbol{m}}, \boldsymbol{v}) I(\boldsymbol{n} - \boldsymbol{m}) \quad (8)$$

where the velocity-dependent filter coefficients $H(\boldsymbol{m}; \acute{\boldsymbol{m}}, \boldsymbol{v})$, in the sample domain may be pre-computed using

$$H(\boldsymbol{m}; \acute{\boldsymbol{m}}, \boldsymbol{v}) = \sum_{k_y=-B_y}^{+B_y} \sum_{k_x=-B_x}^{+B_x} G^*\left(\acute{\boldsymbol{m}}; \frac{k_x}{M_x}, \frac{k_y}{M_y}, \boldsymbol{v}\right) G\left(\boldsymbol{m}; \frac{k_x}{M_x}, \frac{k_y}{M_y}, \boldsymbol{v}\right) \quad (9)$$

or after combination of the $G$ terms

$$H(\boldsymbol{m}; \acute{\boldsymbol{m}}, \boldsymbol{v}) = \frac{1}{M_x M_y M_z} \sum_{k_y=-B_y}^{+B_y} \sum_{k_x=-B_x}^{+B_x} G\left(\boldsymbol{m} - \acute{\boldsymbol{m}}; \frac{k_x}{M_x}, \frac{k_y}{M_y}, \boldsymbol{v}\right). \quad (10)$$

Summations over frequency indices in Eq. 10 may be eliminated using the relationship

$$\sum_{k=-B}^{+B} e^{j2\pi \frac{k}{M}[m - \acute{m}]} = \frac{\sin(\pi W [m - \acute{m}]/M)}{\sin(\pi [m - \acute{m}]/M)} \quad (11)$$

to yield the following closed-form expression for the sample-domain background-enhancing filter coefficients:

$$H(\boldsymbol{m}; \acute{\boldsymbol{m}}, \boldsymbol{v}) = \frac{W_x W_y}{M_x M_y M_z} \times$$

$$\mathcal{D}_{W_x}([m_x - \acute{m}_x - v_x(m_z - \acute{m}_z)]/M_x) \times$$
$$\mathcal{D}_{W_y}\left(\left[m_y - \acute{m}_y - v_y(m_z - \acute{m}_z)\right]/M_y\right) \tag{12}$$
where $\mathcal{D}_W$ is the Dirichlet kernel of order $W$, or periodic sinc function, defined here as
$$\mathcal{D}_A(a) = \frac{\sin(\pi A a)}{A\sin(\pi a)} \tag{13}$$
arising from the 'symmetric sum' of $A$ sinusoids in either the sample or frequency domains. It has $A - 1$ nodes on the interval $a = [0,1]$ and it is normalized to give a maximum limiting value of unity as $a$ approaches zero. Additionally, $\mathcal{D}_A$ has periodic symmetry such that for any integer $\alpha$: $\mathcal{D}_A(a \pm \alpha) = \mathcal{D}_A(a)$ when $A$ is odd and $\mathcal{D}_A(a \pm \alpha) = (-1)^\alpha \mathcal{D}_A(a)$ when $A$ is even. For integer-valued $\acute{\boldsymbol{m}}$, the low-pass background-enhancing filter may be converted to a high-pass prediction-error filter (PEF), to suppress the background signal and enhance the foreground signal (if any), using
$$J(\boldsymbol{n} - \acute{\boldsymbol{m}}) = I(\boldsymbol{n} - \acute{\boldsymbol{m}}) - \hat{I}(\boldsymbol{n} - \acute{\boldsymbol{m}}) \ . \tag{14}$$
In the absence of modeling errors, i.e. when Eq. 4 holds exactly, the output of the PEF, or the residual $J$, contains white-noise plus foreground-signals that are outside the spatiotemporal band of the background signal, due to different shape/texture or motion.

The background enhancing filter may also be applied in the frequency domain using
$$\hat{I}(\boldsymbol{n} - \acute{\boldsymbol{m}}) = \sum_{k_z=0}^{M_z-1} \sum_{k_y=-B_y}^{+B_y} \sum_{k_x=-B_x}^{+B_x} \mathcal{H}^*(\boldsymbol{k}; \acute{\boldsymbol{m}}, \boldsymbol{v}) S(\boldsymbol{k}; \boldsymbol{n}) \tag{15a}$$
or alternatively
$$\hat{I}(\boldsymbol{n} - \acute{\boldsymbol{m}}) \cong \sum_{k_z=-B_z}^{+B_z} \sum_{k_y=-B_y}^{+B_y} \sum_{k_x=-B_x}^{+B_x} \mathcal{H}(\boldsymbol{k}; \acute{\boldsymbol{m}}, \boldsymbol{v}) S(\boldsymbol{k}; \boldsymbol{n})$$
$$\cong \sum_{\boldsymbol{k}=-\boldsymbol{B}}^{+\boldsymbol{B}} \mathcal{H}(\boldsymbol{k}; \acute{\boldsymbol{m}}, \boldsymbol{v}) S(\boldsymbol{k}; \boldsymbol{n}) \tag{15b}$$
where $\boldsymbol{k} = [k_x, k_y, k_z]$ and $B_z \leq K_z$ if $M_z$ is odd or $B_z < K_z$ if $M_z$ is even, in cases where the background is known to be a slow-moving low-frequency texture, so that high temporal-frequency content is negligible. In Eq. 15, $\mathcal{H}$ is the discrete Fourier transform (DFT) of $H$ and $S$ is the 'local' DFT of a 3-D data block extracted from $I$, using the $M_x \times M_y \times M_z$ analysis window, at $\boldsymbol{n} = [n_x, n_y, n_z]$. These transformed quantities are found using
$$\mathcal{H}(\boldsymbol{k}; \acute{\boldsymbol{m}}, \boldsymbol{v}) = \sum_{\boldsymbol{m}=0}^{\boldsymbol{M}-1} F^*(\boldsymbol{m}; \boldsymbol{k}) H(\boldsymbol{m}; \acute{\boldsymbol{m}}, \boldsymbol{v}) \tag{16}$$
and
$$S(\boldsymbol{k}; \boldsymbol{n}) = \sum_{\boldsymbol{m}=0}^{\boldsymbol{M}-1} F(\boldsymbol{m}; \boldsymbol{k}) I(\boldsymbol{n} - \boldsymbol{m}) \tag{17}$$
where

$$F(\boldsymbol{m};\boldsymbol{k}) = \frac{1}{\sqrt{M_x M_y M_z}} e^{j2\pi\left(\frac{k_x}{M_x}m_x + \frac{k_y}{M_y}m_y + \frac{k_z}{M_z}m_z\right)} \quad (18)$$

Equation 11 is again used, to yield the following closed-form expression for the frequency-response of the background-enhancing filter:

$$\mathcal{Q}(\boldsymbol{f};\boldsymbol{\acute{m}},\boldsymbol{v}) = \frac{1}{\sqrt{M_x M_y M_z}} \times$$

$$\sum_{k_y=-B_y}^{+B_y} \sum_{k_x=-B_x}^{+B_x} b_x b_y b_z c_x(f_x) c_y(f_y) c_z(f_z) d_x(f_x) d_y(f_y) d_z(f_z) \quad (19a)$$

where

$$b_x = e^{-j2\pi\frac{k_x}{M_x}(\acute{m}_x - \Delta_x)},\ b_y = e^{-j2\pi\frac{k_y}{M_y}(\acute{m}_y - \Delta_y)} \quad (19b)$$

$$b_z = e^{+j2\pi(v_x k_x/M_x + v_y k_y/M_y)(\acute{m}_z - \Delta_z)} \quad (19c)$$

$$c_x(f_x) = e^{-j2\pi f_x \Delta_x},\ c_y(f_y) = e^{-j2\pi f_y \Delta_y},\ c_z(f_z) = e^{-j2\pi f_z \Delta_x} \quad (19d)$$

$$d_x(f_x) = \mathcal{D}_{M_x}(f_x - k_x/M_x),\ d_y(f_y) = \mathcal{D}_{M_y}(f_y - k_y/M_y) \quad (19e)$$

$$d_z(f_z) = \mathcal{D}_{M_z}(f_z + v_x k_x/M_x + v_y k_y/M_y) \quad (19f)$$

The velocity-induced frequency-shift, that tilts the spatial frequencies out of the $xy$ plane (where $f_z = 0$) according to Eq. 5b, results in a plane passing through frequencies in the $z$ dimension that do not necessarily coincide with the discrete frequency bins at $f_z = k_z/M_z$ of the DFT[2,15,19,20]. The Dirichlet kernel of order $M_z$ in the $z$ dimension is therefore required to capture the 'sidelobes' that result when the 'energy' of each sinusoidal $xy$ component 'spills' into adjacent bins due to the misalignment of the nodes of $\mathcal{D}_{M_z}(f_z)$ with the bins at $f_z = k_z/M_z$. In contrast, the nodes of $\mathcal{D}_{M_x}(f_x)$ and $\mathcal{D}_{M_y}(f_y)$ *do* coincide with the DFT bins at $f_x = k_x/M_x$ and $f_y = k_y/M_y$, therefore the Dirichlet kernel is only used to interpolate the filter response in between the DFT bins in the spatial dimensions. The $b$ factors in Eq. 19 perform the synthesis operation; the $c$ factors and the $\Delta_{xyz}$ constants compensate for the displacement of the sample-domain origin from the centre of the analysis window – as 'displacement' in the sample domain is 'modulation' in the frequency domain. For analysis windows with edge-referenced delay indexing, i.e. $m = 0 \ldots M-1$, for odd or even $M$, the required modulation is found using $\Delta = (M-1)/2$ for each dimension in Eq. 19d. These factors are not required if a non-causal filter is used with centre-referenced delay indexing, i.e. $m = -K \ldots +K$ for odd $M$, which is a feasible option for the spatial dimensions. The frequency-domain filter coefficients $\mathcal{H}(\boldsymbol{k};\boldsymbol{\acute{m}},\boldsymbol{v})$, are found by evaluating $Q(\boldsymbol{f};\boldsymbol{\acute{m}},\boldsymbol{v})$ at the design frequencies, i.e. the bins of the DFT, where $\boldsymbol{f} = [k_x/M_x\,, k_y/M_y\,, k_z/M_z]$. For windows of odd length in all dimensions with centre-referenced delay indexing and $\boldsymbol{\acute{m}} = 0$, the $b$ and $c$ factors are all equal to unity; furthermore, due to the nodal structure of $\mathcal{D}$, the (real) filter coefficients of the resulting linear-phase filter may simply be found using

$$\mathcal{H}(\boldsymbol{k};\boldsymbol{\acute{m}},\boldsymbol{v}) = \frac{1}{\sqrt{M_x M_y M_z}}\mathcal{D}_{M_z}\left(k_z/M_z + v_x k_x/M_x + v_y k_y/M_y\right) \tag{20}$$

for $-B \le k \le +B$ in each dimension. The extra complication associated with the use of analysis windows of even length with edge indexing is necessary to accommodate standard base-two FFT implementations. Furthermore, as discussed in Sec. 3, synthesis at multiple non-central samples allows a greater rate of data throughput because more filtered samples are produced for each FFT-processed block. The loss of phase linearity associated with this approach can be controlled to yield an appropriate balance between processing speed and filter performance (see Sec. 5). For large $M$, and $\acute{m}$ close to $M/2$, the deviation is small. The nodes of the Dirichlet kernels in Eq. 19 result in a very 'lumpy' frequency response. Application of a tapered window via a multiplication in the sample domain helps to 'smooth out' the response via a convolution in the frequency domain. The presented approach may be regarded as a crude frequency-sampling filter-design method, commonly used to design 1-D and 2-D filters. This simple approach was adopted to help offset the extra complexity associated with the extension to 3-D and the unusual geometry of the pass-band – ideally, a tilted plane of finite thickness. The use of a frequency *continuum* in the pass-band instead of a set of discrete frequency *points* during the design phase results in: *integrals* instead of *summations* in the frequency domain and *sinc-function* products instead of *Dirichlet-kernel* products in the sample domain. After the sinc functions are truncated by the analysis window in the sample domain, the ideal pass-band of the filter is convolved with the Dirichlet kernel, which causes the *actual* response to deviate from the *desired* response. The application of a tapered window (in either the sample domain or the frequency domain) reduces the side-lobe level in the actual response. To avoid these issues, an optimal procedure – using an equi-ripple or minimal integral-squared-error criterion, for example – could be used instead.

Finally, as in the sample-domain case, the output of the frequency-domain background-subtraction filter is found using the output of the foreground-enhancing filter in Eq. 14.

### 2.2 Velocity Estimation

So far it has been assumed that the velocity of the background is known *a priori*; although in most cases it is safe to assume that it will need to be estimated. Any number of well-established optical-flow techniques could be used for this purpose, such as gradient-based[21,22,23,24], phase-based[25,26], 2-D block-matching methods[27] or other frequency-domain methods such as those involving Gaussian derivative filters[28] or the complex lapped transform[29]. A 3-D block-matching method is used here because it is conceptually and architecturally compatible with the filtering approach presented in the previous Subsection. In the same way that 2-D block-matching methods use local blocks from consecutive frames to generate the 2-D *cross*-correlation function, the 3-D method employed here uses a local 3-D block of data to generate the 3-D *auto*-correlation function. This is computed most efficiently in the frequency domain, which is the main reason why a frequency-domain block-based approach to filtering is also adopted.

The power spectrum of the image data within the 3-D analysis window at $\boldsymbol{n}$ is

$$P(\boldsymbol{k};\boldsymbol{n}) = S^*(\boldsymbol{k};\boldsymbol{n})S(\boldsymbol{k};\boldsymbol{n}) \tag{23}$$

and the auto-correlation function of the windowed data, with a periodic boundary condition, is

$$R(\boldsymbol{l};\boldsymbol{n}) = \sqrt{M_x M_y M_z}\sum_{k_z=0}^{k_z=M_z-1}\sum_{k_y=-B_y}^{k_y=+B_y}\sum_{k_x=-B_x}^{k_x=+B_x} F^*(\boldsymbol{l},\boldsymbol{k})P(\boldsymbol{k};\boldsymbol{n}) \tag{24}$$

for $-L \le l \le +L$, where $L = M - 1$ (i.e. the maximum 'measureable' displacement given the data block dimensions), $\boldsymbol{l} = [l_x, l_y, l_z]$ is the index vector of sample displacements and $R(\boldsymbol{0}) = \sum_{\boldsymbol{k}\in\boldsymbol{K}} P(\boldsymbol{k};\boldsymbol{n}) = \sum_{\boldsymbol{m}\in\boldsymbol{M}} I(\boldsymbol{n}-\boldsymbol{m})^2$ is the total image power for all samples within the analysis window. Interpolating displacements are readily computed for non integer $l$ values. Note that phase information is lost when the (real) power spectrum ($P$) is created from the (complex) frequency spectrum ($S$); as a consequence, fine spatiotemporal detail is not preserved in $R$. When interpreting $R$, 3-D displacements are converted to velocities using $v_x = l_x/l_z$ and $v_y = l_y/l_z$; thus the auto-correlation 'slice' $R_z$ at $l_z = 1$ gives a coarse indication of motion up to the maximum velocity that may be 'measured' using a data block of the specified size, while slices closer to $L_z$ give a finer indication of motion over a smaller velocity range. In any given slice, the velocity estimate is derived from the combination of $x$-$y$ displacements, for which $R_z$ is maximized. False artifacts due to the assumed periodic-boundary condition are minimized if $M$ is large and $l$ is small, i.e. using $L_{xyz} \ll M_{xyz} - 1$; alternatively, zero-padding could be used at an extra computational cost[29].

Using this approach to derive velocity information from the 3-D *auto*-correlation function of spatiotemporal data *blocks*, as an alternative to the more conventional technique involving the 2-D *cross*-correlation function of (consecutive) spatial data *frames*, gives greater noise immunity due to time 'averaging'. Use of the real-valued auto-correlation function is ideal for background analysis because it neglects phase information which 'defocuses' the filter so that it considers *average* motion throughout the whole analysis window, as opposed to *specific* motion concentrated at points within the window, which would be the case if the power output of each velocity-tuned filter is analyzed. Other background velocity-estimation methods were also considered, such as the fitting of planes to $P$ in the frequency domain[19] , or the fitting of lines to $R$ in the sample domain; however the method described in this Subsection was chosen for its simplicity, reliability and speed. To avoid velocity discretization, on-the-fly design of finely tuned filters using Eq. 12 or Eq. 19 in an iterative optimization procedure may also be appropriate in applications where estimation accuracy is more important than execution speed[20].

## 3 Implementation

The input image is partitioned into overlapping analysis blocks – with the length in each dimension ($M_{xyz}$) equal to an integer power of two – and the spectrum of the data block is generated efficiently using the FFT. A smaller synthesis block – with lengths in each dimension ($\acute{M}_{xyz} = 2\acute{K}_{xyz}$) equal to an even number of samples – is defined within each analysis block. The analysis and synthesis blocks are concentric and the overlap of the analysis blocks is set so that adjacent synthesis blocks abut. Application of a *local* FFT allows non-uniform motion to be accommodated; use of a synthesis *block* (rather than a *sample*) means that the FFT of a single data block may be reused to filter multiple samples; using $\acute{K}_{xyz} < K_{xyz}$ improves performance by excluding samples near the edge of the analysis window, where phase non-linearity and magnitude variability is greatest. Prediction errors are large for signals with components that are midway between analysis frequency bins and the errors increase with the distance of the synthesis sample from the centre of the analysis window. Use of tapered window functions reduces the error by broadening the response of each frequency bin, thus the frequency selectivity of the filter. Prediction out to the edge of the analysis window is necessary for analysis blocks around the perimeter of the image frame if processed outputs are required for all pixels. The proposed filter framework with $\acute{K}_{xyz} > K_{xyz}$, potentially offers a solution to the problem of edge-artifact mitigation in image filtering[30]; however, this aspect of the problem was not considered here.

In most applications, the motion of the background is non-uniform, time varying, and not known *a priori*. The local velocity estimate $\hat{\boldsymbol{v}}$, was computed using the 3-D auto-correlation function of a given data block. Only the $l_z = 1$ slice was constructed, with velocity hypotheses $v_{xy} = \acute{l}_{xy}/\acute{L}_z$ for integer $-\acute{L}_{xy} \leq \acute{l}_{xy} \leq +\acute{L}_{xy}$. The acute accent is used for these analysis variables to indicate that they are not necessarily derived from the analysis window dimensions; instead, they are arbitrarily selected to give the desired velocity grid extent and density. Filters were pre-designed for each velocity hypothesis on the grid; their coefficients were computed using Eq. 19 on start-up and stored for later re-use at run-time, although Eq. 12 and Eq. 16 could also have been used for the same result. The filter corresponding to the velocity of the auto-correlation maximum in a given data block was used to estimate the background intensity at the synthesis sample using Eq. 15; the associated prediction error was then computed using Eq. 14.

Complex notation has been used for convenience in the previous Section; however, it should be noted that the input ($I$) and most of the outputs ($H$, $P$ and $R$, with the exception of $\mathcal{H}$) are real-valued. Real (single-precision) data types were used in the C software implementation to represent real and imaginary parts of complex numbers.

## 4 Simulation

Synthetic translating and diverging backgrounds were generated to simulate scenes measured by downward- and forward-looking infrared sensors mounted on a fast non-maneuvering aircraft. Ten randomly instantiated data sets with $N_{xyz} = 64$ were produced for each scenario.

### *4.1 Translating Background*

Background textures were generated using 16 equally-weighted sinusoidal components, with random frequencies $f_{xy}$ uniformly distributed on the interval $[-B_{xy}, +B_{xy}]/M_{xy}$ cycles per sample (with $M_{xy} = 16$ and $B_{xy} = 3$) and random 'relative' phase offsets uniformly distributed on the interval $[0,2\pi]$ radians. The frequency of each component, as a function of $n_z$, was shifted using velocity components ($v_{xy}$) uniformly distributed on the interval $[-2, +2]$ pixels per frame. Only the real part of the complex input signal was used. The intensity of each instantiation was normalized to yield zero average amplitude and unity average power. Four *scenarios*, which are *variants* of the Translating background *type*, were created: Unmodified (TU), Low-power noise added (TL), High-power noise added (TH) and Foreground-target injected (TF). The noise was drawn from a zero-mean uncorrelated Gaussian distribution with a variance selected to yield background signal-to-noise-ratios (SNRs) of 20 dB and 10 dB in the TL and TH scenarios, respectively. The foreground point-target variant of TF moved in a circular orbit around the centre of the field of view (FOV) with a constant radius of 12 pixels and a tangential speed uniformly distributed on the interval $[-2, +2]$ pixels per frame and a starting angle uniformly distributed on the interval $[0, 2\pi]$ radians. A Gaussian point-spread function (PSF) was used to mix the point target with the background. The PSF had a standard deviation of 1 pixel and a 'hard' cut-off of 2 pixels.

*4.2 Diverging Background*

This background was generated by applying a bank of 3-D background-enhancing filters to a random-noise input-sequence. The background moved in a tangential direction, relative to the centre of the FOV with the speed at each pixel determined using $\left(4R\left[N_{xy}-1\right]\right)/\left(2R^2+\left[N_{xy}-1\right]^2\right)$ where $R$ is the distance (in pixels) from the FOV centre at $\left(N_{xy}-1\right)/2$, giving a maximum velocity of $v = [\pm 1, \pm 1]$ and a speed of $\sqrt{2}$ pixels per frame at the corners of the FOV. A bank of 64 x 64 unique filters was therefore designed using the background velocity at each of the pixels in the FOV. The coefficients of the linear-phase background-generating filters for $m_{xyz} = -K_{xyz} \ldots +K_{xyz}$ were computed using Eq. 12 with $K_{xyz} = 8$ (to give $M_{xyz} = 17$), $B_{xy} = 3$ (to give $W_{xy} = 7$) and $\acute{m} = 0$. The zero-mean Gaussian-noise input was extended in both directions of each dimension by $K_{xyz}$ samples to yield a filtered output with the desired dimensions. After normalization, a point target with the same random trajectory parameters as the TF scenario was also inserted into the diverging scene, to create the Diverging background with Foreground-target injected (DF) scenario; however, its PSF had a standard deviation of $1/2$ a pixel, and a hard cut-off of 1 pixel.

*4.3 Filters*

Multiple variants of three basic filter types were also created and used to process the synthetic data. The basic filter types are: the proposed 3-D filter, a more conventional 2-D filter and a standard Lucas-Kanade gradient-based optical-flow filter[21,22,23]. The following 3-D filters were designed: a Large Analysis window filter, optimized for the fast-moving low-frequency uniformly Translating background with a diffuse foreground target (3D/LAT); a similar filter with a Smaller Analysis window for greater execution speed (3D/SAT); and a filter optimized for the non-uniform and less-coherent motion of the Diverging background with a more concentrated foreground target (3D/DIV). Analogous 2-D filters were designed. The first was the 2-D equivalent of the 3D/LAT filter (2D/LAT); the second had a Finer Velocity Grid (2D/LAT/FVG), which was made feasible by the increased execution speed of 2-D filters, relative to their 3-D counterparts. Coarse and fine versions of the 3-D/DIV filter were also designed (2D/DIV & 2D/DIV/FVG). The 2-D filters whitened the background using only spatial information in the current frame and used the cross-correlation between the current and the previous frame to estimate velocity (i.e. correlation-based or block-matched optical-flow). The 2-D filters, which do not employ time integration, were used to demonstrate the performance gain (if any) brought about by joint spatiotemporal processing. The Lucas-Kanade optical-flow algorithm does, on the other hand, implicitly utilize multi-frame information through the use of numerical Derivatives, computed using independent $M$-point central-difference operators in the temporal and spatial dimensions, applied to consecutive spatially low-pass filtered frames, then followed by the summation of gradients over a local spatial window and the least-squares solution of the optical-flow equations. This (LKD) filter was used for the purpose of velocity-field accuracy comparison only, as it does not output a whitened image.

The aforementioned filters were designed using the parameters defined in Table 1 and the main characteristics of each filter are summarized below. The parameters were chosen with both execution speed and estimation accuracy in mind. Average processing rates (in seconds per frame), for C code running on a personal computer with a T9400 central processing unit, are also given below in parentheses.

a) **3D/SAT** (0.044): A ‘fast’ 3-D filter for the translating background scenario.
b) **3D/LAT** (0.063): Same velocity grid as above, with larger analysis and synthesis windows and the filter bandwidth approximately maintained.
c) **2D/LAT** (0.0089): A 2-D version of the above filter.
d) **2D/LAT/FVG** (0.024): Same as above, with a finer velocity grid.
e) **3D/DIV** (0.022): A 3-D filter optimized for the diverging background scenario.
f) **2D/DIV** (0.0075): A 2-D version of the above filter.

g) **2D/DIV/FVG** (0.017): Same as above, with a finer velocity grid.

h) **LKD** (0.0029): All low-pass filtration, intensity derivative computation, and local derivative summation, operations were performed over windows with $M = 5$. The Gaussian convolution kernel of the 2-D low-pass 'blur' filter had a standard deviation of 1 pixel.

TABLE 1

PARAMETERS FOR THE 2-D AND 3-D FILTERS

| Filter | $M_{xy}$ | $M_z$ | $\acute{M}_{xy}$ | $\acute{M}_z$ | $B_{xy}$ | $B_z$ | $\acute{L}_{xy}$ | $\acute{L}_z$ |
|---|---|---|---|---|---|---|---|---|
| 3D/SAT | 16 | 8 | 4 | 2 | 3 | 4 | 8 | 4 |
| 3D/LAT | 32 | 16 | 8 | 2 | 6 | 8 | 8 | 4 |
| 2D/LAT | 32 | 1 | 8 | 1 | 6 | 0 | 8 | 4 |
| 2D/LAT/FVG | 32 | 1 | 8 | 1 | 6 | 0 | 16 | 8 |
| 3D/DIV | 16 | 8 | 4 | 2 | 3 | 4 | 4 | 4 |
| 2D/DIV | 16 | 1 | 4 | 1 | 3 | 0 | 4 | 4 |
| 2D/DIV/FVG | 16 | 1 | 4 | 1 | 3 | 0 | 8 | 8 |

The code for all algorithms was not optimized for speed and all implementations should be regarded as experimental prototypes. None of the algorithms made use of recursive computation, for example: sliding frequency analysis in the 2-D and 3-D filters, or running sums in the LKD filters. All code was executed in a single thread.

The LKD filter is clearly the fastest filter. If the data processing rate had been computed on a per pixel basis, the speed gap would open even further, as it processes more pixels per frame due to its small analysis-window size, which does not leave such a large margin of unprocessed pixels around the perimeter of each frame. The 3-D filters are several times slower than their 2D counterparts. Both filter types may be accelerated by:

1) Increasing the analysis window size ($M$), so fewer but larger FFTs are applied;
2) Decreasing the velocity grid extent (using $\acute{L}_{xy}$) and density (using $\acute{L}_z$);
3) Increasing the synthesis window size ($\acute{M}$) and/or
4) Decreasing the filter bandwidth ($f_{\mathrm{BW}} = 2B/M$).

The speed/performance tradeoff may be summarized as follows: The first alternative reduces the ability of the filter to handle non-uniform motion; the second, decreases the velocity estimate accuracy; the third approach increases the 'granularity' of the output and introduces block artefacts; whereas the fourth option means that high-frequency features in the background may not be whitened properly (if fast-moving high-frequency components are expected then a high temporal bandwidth is required in the 3-D filter, see Eq. 5b). The selection of the 3-D filter parameters required a few iterations before suitable combinations were found, although the equations in Section 2 may be used as a guide.

### *4.4 Metrics*

The primary purpose of the filters is to enhance the foreground/background contrast by attenuating the background signal; a background velocity estimate at each pixel (or block) is a secondary output that may also be used to enhance 'downstream' target tracking and image understanding functions. The performance of these downstream functions was not specifically examined here. To quantify filter performance, the root-mean-squared (RMS) velocity error of the background motion-field was computed for all scenarios and the signal-to-clutter ratio (SCR) was computed for scenarios with a target in the foreground (TF & DF). The SCR (on a dB scale) for a given data set was calculated by dividing the average power within a 2x2 pixel *region* centered on the true target position in every frame by the average power of *all* pixels in the data set (which is assumed to be dominated by the background signal). Filters that whiten the background without attenuating the foreground have a large SCR. Aggregate SCR and RMS metrics are presented in Table 2 and Table 3, respectively. Aggregate SCRs were computed by averaging the linear ratios over all data sets.

TABLE 2

AGGREGATE RMS VELOCITY ERROR (PIXELS PER FRAME)

FOR ALL SCENARIOS AND THE THREE FILTER TYPES

| Filter Type | TU | TL | TH | TF | DF |
|---|---|---|---|---|---|
| 3-D | 0.17 [a] | 0.17 [a] | 0.17 [a] | 0.18 [a] | 0.13 [e] |
| | 0.11 [b] | 0.11 [b] | 0.11 [b] | 0.12 [b] | - |
| 2-D | 0.11 [c] | 0.11 [c] | 0.11 [c] | 0.12 [c] | 0.15 [f] |
| | 0.07 [d] | 0.07 [d] | 0.08 [d] | 0.08 [d] | 0.14 [g] |
| LKD | 0.26 | 0.26 | 0.32 | 0.31 | 0.22 |

TABLE 3

AGGREGATE SIGNAL-TO-CLUTTER RATIOS (ON A dB SCALE)

FOR TWO SCENARIOS AND TWO FILTER TYPES

| Filter Type | TF | DF |
|---|---|---|
| Raw Data | 5.26 | 3.21 |
| 3-D | 20.49 [a] | 7.58 [e] |
| | 22.24 [b] | - |
| 2-D | 15.43 [c] | 8.28 [f] |
| | 15.43 [d] | 8.28 [g] |

Filtered using: [a] 3D/SAT, [b] 3D/LAT, [c] 2D/LAT, [d] 2D/LAT/FVG, [e] 3D/DIV, [f] 2D/DIV, [g] 2D/DIV/FVG.

### *4.5 Analysis of Results*

The results presented in Table 2 suggest that that the joint consideration of three dimensions significantly improves the ability of a whitening filter to separate the foreground target from the translating background. The aggregate SCR for the 3-D filter with the large analysis window (3D/LAT) is more than 6 dB greater than the 2-D filters. Use of the smaller analysis window (3D/SAT) also results in a net improvement; however the impact is somewhat reduced (a little over 4 dB). Closer analysis of the individual results confirmed that the enhancement is most pronounced for large foreground/background velocity differences. Example data from one of these cases are displayed for the 3-D in Figure 1. The 3-D whitening filter clearly enhances the target visibility; with a significantly increased foreground/background contrast.

The simulation results confirm that the 3-D design has the intended effect for the *translating* background. However the whitening performance of the 3-D filter (3D/DIV) is slightly worse than the corresponding 2-D filter (2D/DIV) for the *diverging* background (see Table 3 and Figure 2). This is probably due to the evolving nature of the scene. Unlike the translating background, the 'blob'-like features in the diverging background appear, disappear and slowly change their shape over time. Thus the use of long-term spatiotemporal 'correlation' has the potential to provide false 'cues' that 'mislead' the filter; however, the aggregate velocity accuracy of the 3-D filter for this background is greater than all other filters examined (see Table 2).

In Table 2 it can be seen that increasing the filter dimension from two to three (i.e. for 2D/LAT and 3D/LAT filters) has no significant impact on the velocity error for the translating backgrounds. It is also apparent that decreasing the size of the 3-D analysis window (as used in the 3D/SAT filter) does degrade accuracy and that a finer velocity grid (as used in the 2D/LAT/FVG filter) does improve accuracy. Note that the fine velocity-grid of the 2D/LAT/FVG filter has no impact on its whitening performance because 2-D background prediction filter only uses spatial information. The fine velocity grid was not used for the 3-D filter because it would have made the filter too slow. The velocity accuracy of the 2-D and 3-D filters is largely independent of the additive noise power and the presence/absence of the foreground target – the same cannot be said of the LKD filter. The LKD filter's aggregate velocity error is greater than all the other filters in all scenarios; furthermore the error increases with the noise level. Analysis of the errors in individual scenarios revealed the well-known speed dependence of the LKD velocity error. In contrast, the velocity estimation performance of the 2-D and 3-D filters is largely independent the background speed, provided the size of the analysis window and the coverage of the velocity hypothesis grid are sufficient, which is the case in these simulations. Note that the 2-D/3-D filters examined here, and the LKD filter, approach the problem of target detection in different ways – the former filters use *intensity contrast* to support target detection; while the latter filter offers the use of *velocity disparity* as an alternative (see Figure 3).

## 5 Discussion

Clearly there are a large number of possible design permutations here, especially when it is appreciated that the process of frequency estimation in each dimensions is separable and that a different approach may be adopted in each dimension[31]; however, this paper deals with only one approach, which is arguably the simplest from a conceptual perspective. A direct-digital-design approach is adopted to avoid the need for *s*-plane analysis and unforeseen artifacts associated with the discretization of an analog prototype[9,10,11]. Like the approach taken in Ref. 17, the proposed filter has a finite impulse response in each dimension and is implemented non-recursively; however, the filters used here are far from optimal in a mathematical sense. Block convolution (e.g. overlap and add/save) and recursive (FIR and IIR) filter realizations are currently under investigation and will be reported in the near future. It is possible that using banks of recursive frequency analyzers may outperform the block FFT method employed in this paper, using IIR filters in the temporal dimension, to avoid rounding error accumulation, and FIR filters in the spatial dimensions. Furthermore, recursive approaches will be free of block artefacts in the output. Tapered window functions were not applied to reduce the side-lobes of the frequency response, mainly to avoid the introduction of further design variables.

Figure 4 was constructed to understand how the filter design and implementation choices described in Sec. 2 & Sec. 3, affected the foreground-to-background enhancement performance described in Sec. 4. A 3D/SAT filter tuned to $v_x = 1$ and $v_y = 0$ was designed for various synthesis samples $\acute{m}_{xy} = 8,9,11\,\&\,13$ and $\acute{m}_z = 4$. The theoretical gain of the PEF for foreground and background input-signals was then generated using Eq. 12, as a function of velocity mismatch. Gain as a function of filter/input *angle* mismatch is shown in the upper subplot while gain as a function of filter/input *speed* mismatch in shown in the lower subplot. Ideally, the PEF should strongly attenuate the clutter signal and have unity gain (i.e. 0 dB) for the target signal over a wide range of geometries (i.e. be tolerant of mismatch).

The 3D/SAT filter was designed using *discrete* spatial components with frequencies at $f_{xy} = k_{xy}/M_{xy}$ for $k_{xy} = -B_{xy} \dots +B_{xy}$ (where $M_{xy} = 16$ and $B_{xy} = 3$) yielding a translating pulse-like Dirichlet kernel (with a *periodic* boundary condition) in each 2-D time slice of the impulse response; however, the background (clutter) signal is modeled here using a frequency *continuum* (with coherent phase) over the same interval $-B_{xy}/M_{xy} \le f_{xy} \le +B_{xy}/M_{xy}$ yielding a translating pulse-like sinc function (with a *non-periodic* boundary condition) in each 2-D time slice of the input image. The foreground (target) signal is modeled using a wider bandwidth $-4/M_{xy} \le f_{xy} \le +4/M_{xy}$, to yield a more spatially concentrated pulse.

Using $M_{xy} = 16$ and $\acute{M}_{xy} = 4$ in the filter means that for each block processed for $m_{xy} = 0 \dots 15$ via the FFT, filtered outputs are produced at $\acute{m}_{xy} = 4 \dots 11$, using edge-referenced indexing. The even filters only have a linear-phase response when the synthesis point is at the centre of the analysis window at $\acute{m}_{xy} = (M_{xy} - 1)/2 = 7.5$. Phase non-linearity and magnitude distortion increase as the synthesis sample moves away from this central point – especially at frequencies that fall in between the bins of the DFT. This decreases the ability of the PEF to selectively attenuate the background; however, the data throughput increases because more samples are processed for every 3-D FFT. Figure 4 shows that there is negligible difference in the clutter attenuation for $\acute{m}_{xy} = 8\,\&\,9$; for zero velocity mismatch there is approximately a 3 dB performance loss for $\acute{m}_{xy} = 11$ and 15 dB for $\acute{m}_{xy} = 13$, which suggests that using $\acute{M}_{xy} = 4$ yields a near optimal balance between attenuation performance and execution speed.

For $\acute{m}_{xy} = 8$ & 9, when the clutter velocity is perfectly matched to the filter, the background is attenuated by 29 dB; however, if it is assumed that there is a speed mismatch of ¼ pixel per frame or an angle mismatch of 15° due to the resolution of the velocity grid, then the attenuation is closer to 18 dB. For an orthogonal target signal with an angular mismatch of 90° or more (best case) the attenuation is around 1 dB, but for a perfectly velocity-matched target, the attenuation is closer to 8 dB. Using a gain of -18 dB for the clutter and a gain of -1 dB to -8 dB for the target suggests an SCR improvement of 10-17 dB for the 3D/SAT filter, which is consistent with the observed result of a 15 dB improvement, relative to the raw data, for the translating background (see Table 3).

The performance of the proposed approach assumes that the target is point-like, i.e. featureless with a diameter less than two-four pixels in the image plane. As the target grows in size, it is likely to be 'mistaken' for background and whitened due to obscuration of the background in an undersized spatial analysis window and due to insufficient energy in the high-frequency region of the spectrum.

## 6 Application

A stationary infrared camera on a pan-tilt tracking mount was used to observe a distant aircraft set against a cloudy backdrop. The camera has the ability to track a manually designated target, so that it remains near the centre of the field of view (FOV). The FOV of the camera is 128 x 128 pixels. The IR data collected by the camera were post-processed using the proposed 3-D filter with the following parameters: $M_{xy} = 8$, $M_z = 4$, $\acute{M}_{xy} = 4$, $\acute{M}_z = 2$, $B_{xy} = 2$, $\acute{L}_{xy} = 4$, $\acute{L}_z = 2$. These parameters yield a fairly coarse 9 x 9 velocity grid with velocity increments of $1/2$ from -2 to +2 pixels per frame. A relatively small analysis window was also used because the spatial correlation distance was quite short in these data. With these parameters, a data throughput rate of approximately 30.5 frames per second (or 0.0328 seconds per frame) was achieved. The raw input data and the filtered output data, i.e. the output of the 3-D PEF, are shown in Figure 5 for three different cases. Only a 64 x 64 region of interest centered on the midpoint of the camera's FOV and containing the target are shown. The data are not ideal because there is very little long-range spatial structure in the background and all apparent motion in the background is uniform and self-induced, so that it could be handled by other simpler means; however, the data does serve to highlight the intended domain of application and the operation of the filter in less-than perfect conditions.

When the target is set against a locally dim background, the foreground target 'excites' the background subtraction filter, so that the portion of the target's spectrum that overlaps the background's spectrum is attenuated. This effect is not apparent in the simulated data because the target was always set against a bright moving texture which 'focused the attention' of the filter on the background. This effect does result in some loss of target power in the real data; however, elsewhere in the image the background generally experiences a greater loss, so there is still a net enhancement in the foreground-to-background contrast. As a result of processing, the target becomes the brightest pixel in the scene in all three cases shown in Figure 5. However, as forewarned in earlier Sections, there is some residual structure in the background, due to the use of imperfect background models and approximations. The results could possibly have been improved through the application of a tapered window function such as a Slepian or a sum-of-cosines window.

## 7 Conclusion

The simulations indicate that the proposed 3-D filters may be appropriate to enhance foreground/background intensity contrast in scenes where the background is a delocalized low-pass texture (clutter) and the foreground consists of localized features (point targets). Relative foreground/background motion permits 3-D filters to separate foreground/background features, with overlapping spatial frequencies, which would not otherwise be resolvable using a more conventional 2-D whitening filter. The 3-D filter requires the spatial bandwidth of the background to be approximately known *a priori*. The local velocity of the background is estimated using the 3-D auto-correlation function, as a robust alternative to other optical flow methods such as 2-D block-based methods and the Lucas-Kanade gradient-based method. The estimate is then used to tune in a velocity-matched prediction-error filter (PEF) which whitens the background. The foreground is not severely attenuated by the PEF if one or both of the following conditions are approximately satisfied: a significant proportion of the foreground's spatial frequency content lies outside the stop band of the PEF; and/or the foreground and background velocity are significantly different. This addresses the issue of 1-D temporal-filter versus 2-D spatial-filter selection and brings the two approaches together within a coherent theoretical framework. The Dirichlet kernel is used as a convenient and intuitive tool for the design and analysis of odd and even filters, in either the frequency or sample domains. The design and implementation of 3-D band-pass filters is somewhat more difficult than their 1-D equivalents, therefore a simple approach was adopted to facilitate the synthesis of arbitrary motion-sensitive filters with acceptable performance characteristics for the purpose of point-target enhancement in infrared imaging sensors. The proposed approach may also be suitable in other surveillance system that utilize imaging sensors – infrared, optical, radio-frequency (e.g. radar) or acoustic (e.g. active sonar) – where structured backgrounds interfere with the detection of point targets.

**Hugh L. Kennedy** received B.E. and Ph.D. degrees from The University of New South Wales in 1993 and 2000. He is currently a principal engineer in the Defence and Systems Institute at the University of South Australia. Prior to joining the university in late 2010, he worked in industry on the design, development, integration, and maintenance of a variety of different sensor systems – electro-optic, radio-frequency and acoustic.

## Figures

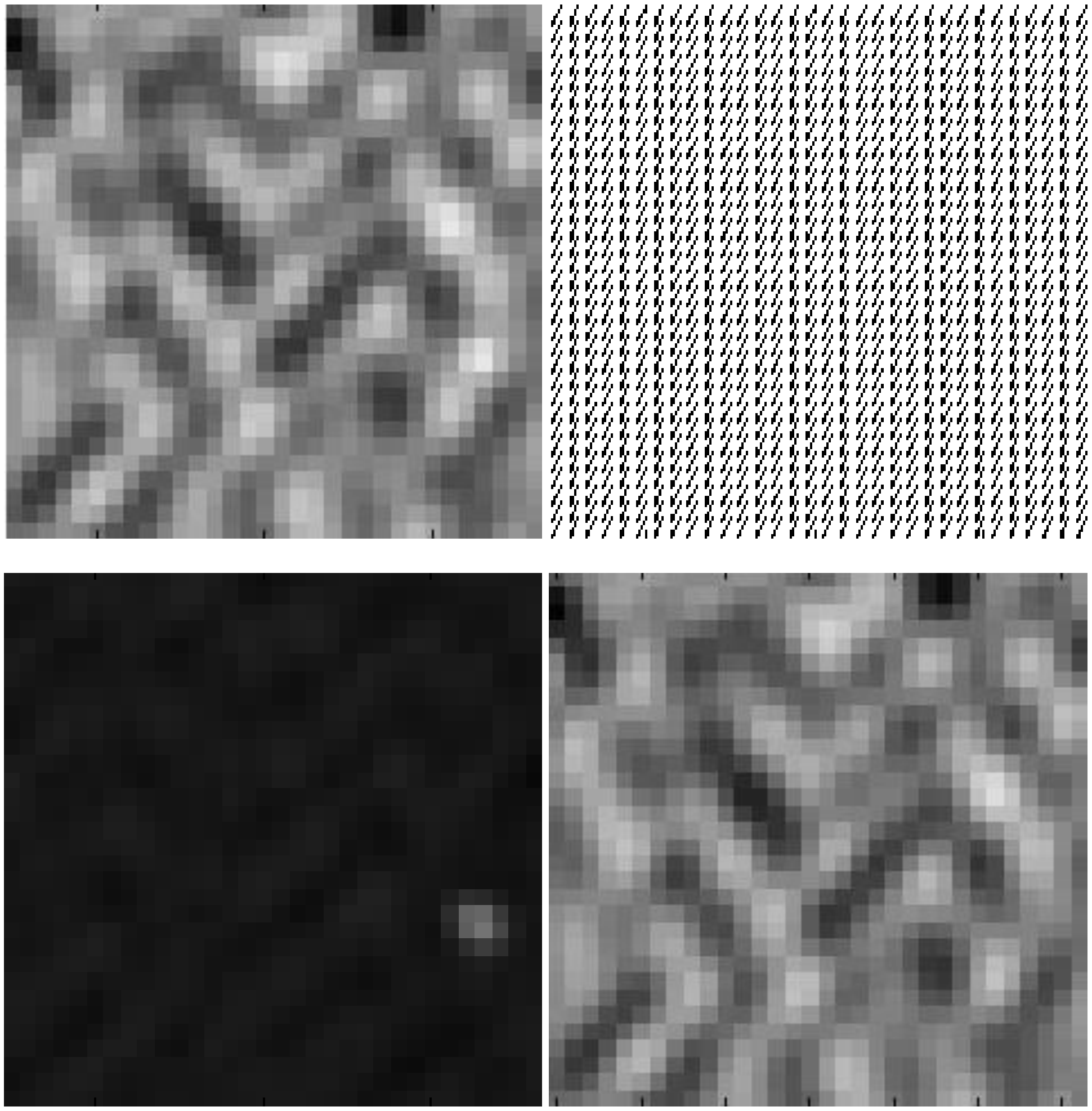

**Fig. 1**. Operation of the 3-D filter (3-D/SAT) for a scene with a translating background and a foreground target. Clockwise from top left: the raw data input, the estimated background velocity field, the predicted background, the whitened output (i.e. the difference between the first and third subplots), where the target is clearly visible near the lower-right corner.

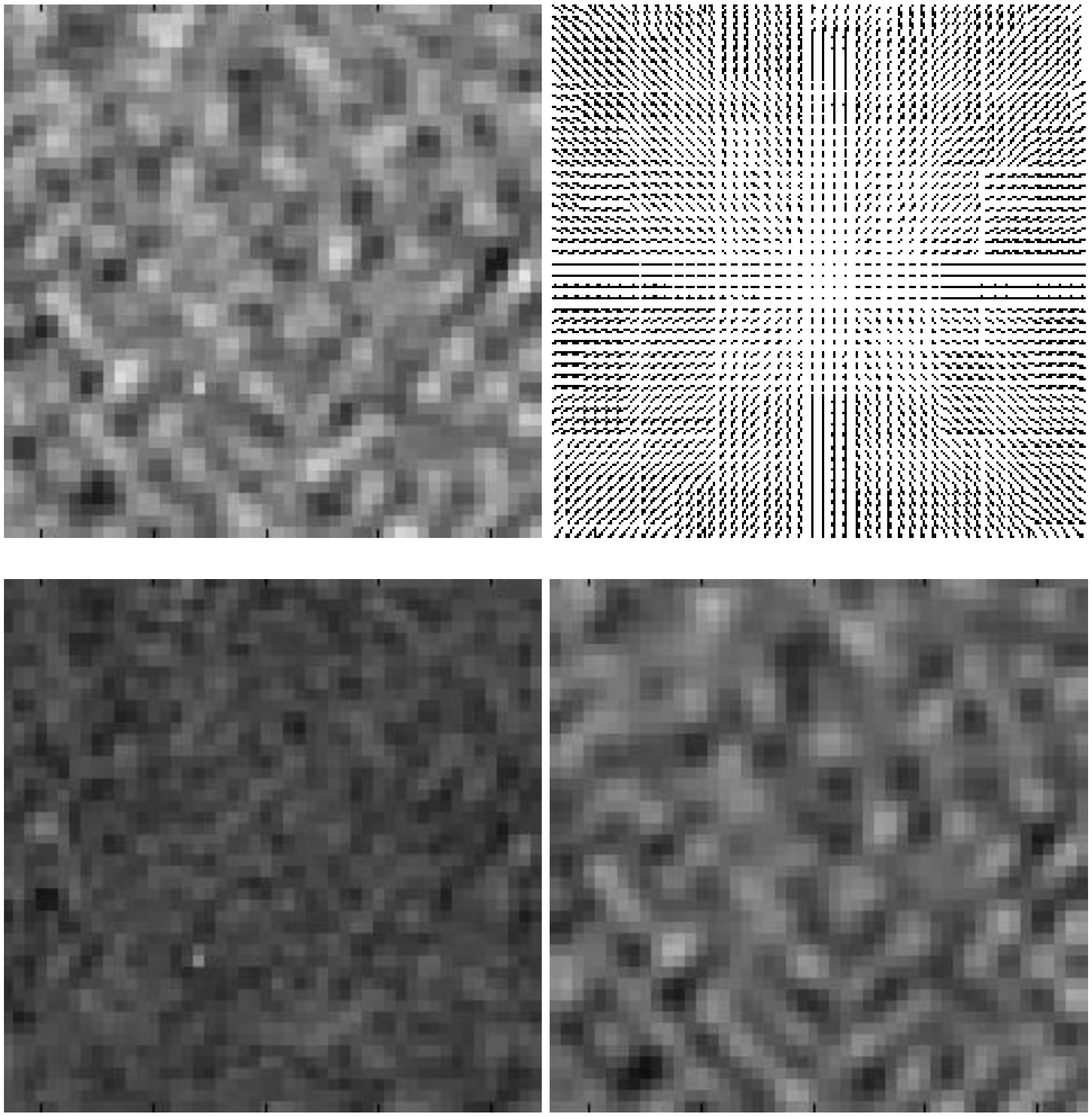

**Fig. 2**. Operation of the 3-D filter (3D/DIV) for a scene with a diverging background and a foreground target. Clockwise from top left: the raw data input, the estimated background velocity field, the predicted background, the whitened output (i.e. the difference between the first and third subplots) – target visible near the bottom left corner.

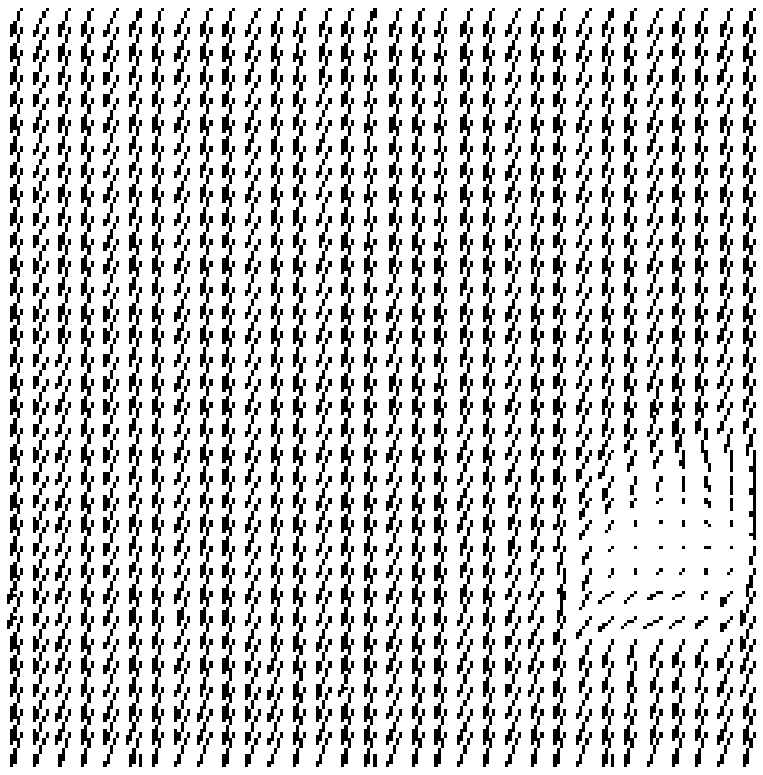

**Fig. 3**. The velocity field of the LKD filter for the translating-plus-target data set depicted in Fig. 1.

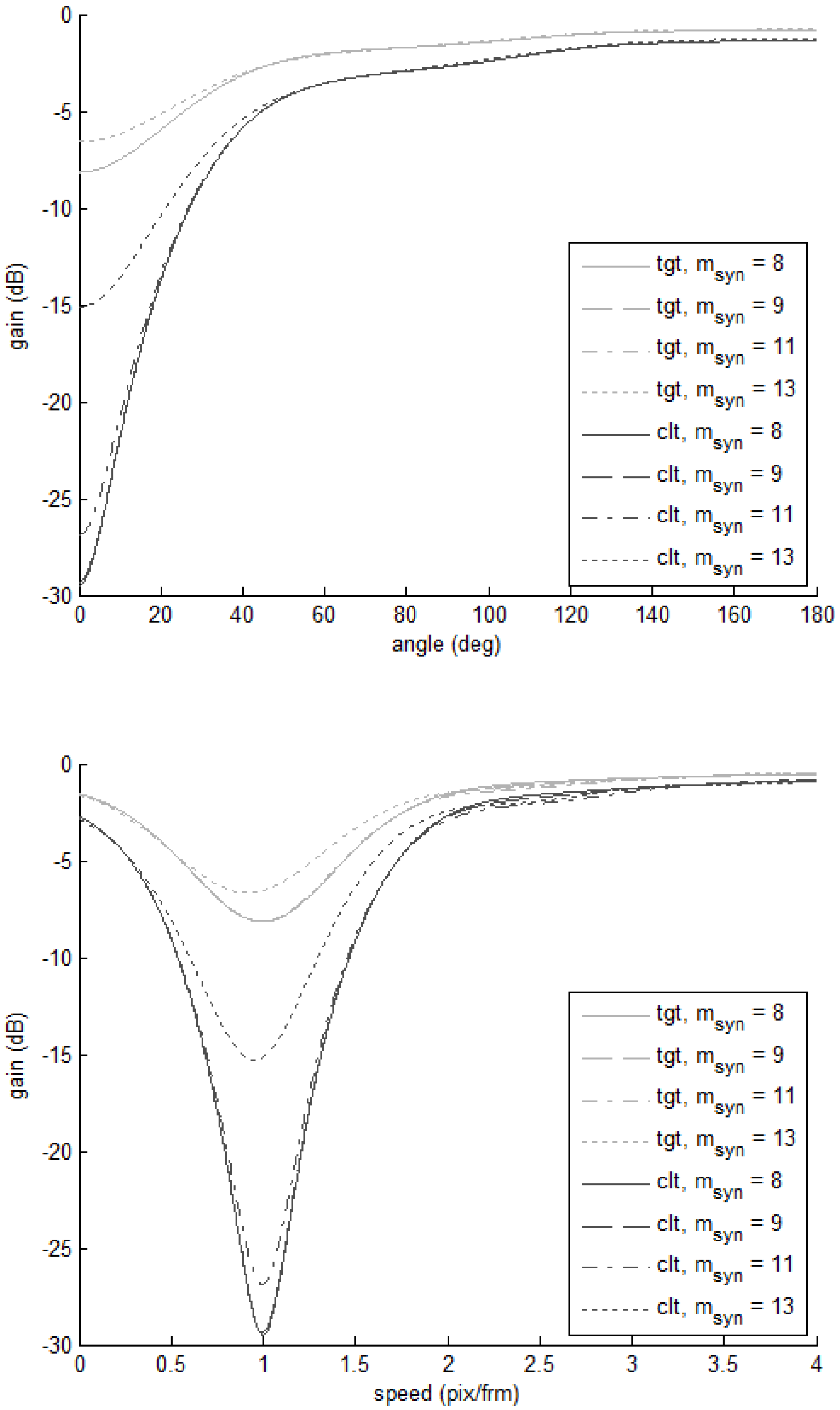

**Fig. 4.** Gain of a prediction-error filter as a function of signal motion direction (top) and speed (bottom) for a foreground target signal (light gray) and a background clutter signal (dark gray), when processed using a (3D/SAT) filter tuned to $v_x = 1$ and $v_y = 0$ for various synthesis sample locations $\acute{m}_{xy}$ within an (edge-referenced) analysis window with $M_{xy} = 16$.

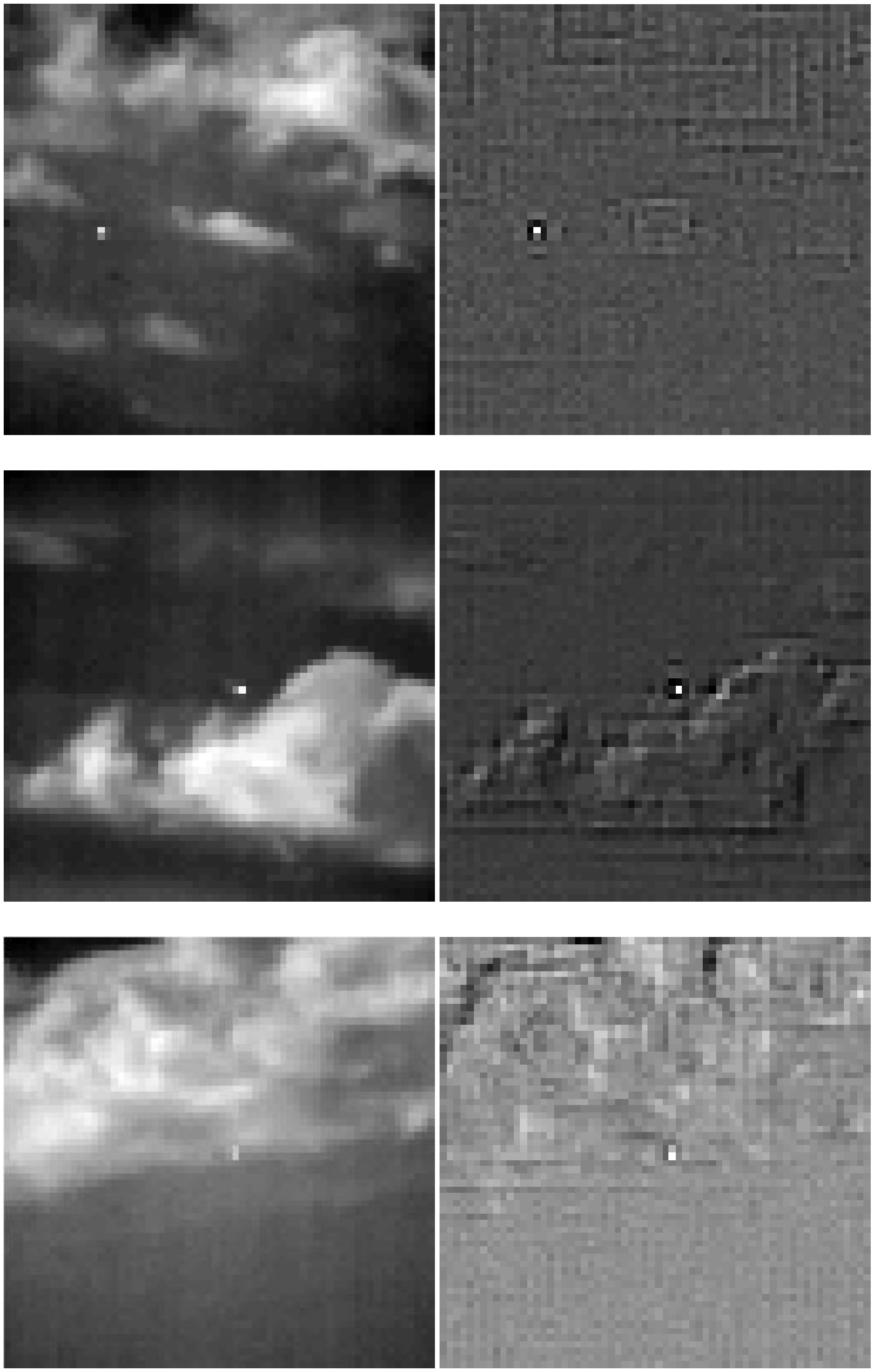

**Fig. 5.** Real infrared data processed using the proposed 3-D filter. Left column: raw input data; Right column: filtered output data. Top row: fixed camera mode; Middle row: tracking camera mode; Bottom row: tracking camera mode.